\documentclass[runningheads]{llncs}
\usepackage[T1]{fontenc}
\usepackage{booktabs}
\usepackage{multirow}
\usepackage{amsfonts}
\usepackage{adjustbox}
\usepackage{amssymb}
\usepackage{xcolor}
\usepackage{float}
\usepackage{subfig}
\usepackage{glossaries}
\usepackage{orcidlink}
\usepackage{xcolor}
\usepackage{hyperref}

\usepackage[flushleft]{threeparttable}

\usepackage{graphicx,verbatim}
\newacronym{llm}{LLMs}{Large Language Models}
\newcommand{\tinycolor}[2]{\textcolor{#1}{\raisebox{0pt}{\footnotesize #2}}}

\begin{document}
\title{Semantic Alignment of Unimodal Medical Text and Vision Representations}
\titlerunning{Semantic Alignment of Unimodal Medical Text and Vision Representations}

\author{Maxime Di Folco\inst{1,2} 
\and Emily Chan \inst{1,2}
\and Marta Hasny \inst{1,2}
\and Cosmin I. Bercea \inst{1,2}
\and Julia A. Schnabel \inst{1,2,3}
}

\authorrunning{M. Di Folco et al.}

\institute{Institute of Machine Learning in Biomedical Imaging, Helmholtz  Munich, Germany
 \and School of Computation, Information and Technology, 
\\ Technical University of Munich, Germany \and
School of Biomedical Engineering \& Imaging Sciences, King's College London, UK\\
}

\maketitle 
\begin{abstract}

General-purpose AI models, particularly those designed for text and vision, demonstrate impressive versatility across a wide range of deep-learning tasks. However, they often underperform in specialised domains like medical imaging, where domain-specific solutions or alternative knowledge transfer approaches are typically required. Recent studies have noted that general-purpose models can exhibit similar latent spaces when processing semantically related data, although this alignment does not occur naturally. Building on this insight, it has been shown that applying a simple transformation - at most affine - estimated from a subset of semantically corresponding samples, known as anchors, enables model stitching across diverse training paradigms, architectures, and modalities. In this paper, we explore how semantic alignment - estimating transformations between anchors - can bridge general-purpose AI with specialised medical knowledge. Using multiple public chest X-ray datasets, we demonstrate that model stitching across model architectures allows general models to integrate domain-specific knowledge without additional training, leading to improved performance on medical tasks. Furthermore, we introduce a novel zero-shot classification approach for unimodal vision encoders that leverages semantic alignment across modalities. Our results show that our method not only outperforms general multimodal models but also approaches the performance levels of fully trained, medical-specific multimodal solutions. Code will be available upon acceptance.



\end{abstract}

\section{Introduction}

With their remarkable adaptability, general-purpose AI models for text and vision excel at extracting meaningful features for a wide range of deep-learning tasks, such as sentiment analysis, question answering and document summarisation for language models \cite{zhao2023surveyllm}, or object detection, image classification and segmentation for vision models \cite{khan2022transformers}. However, when applied to domain-specific tasks, such as the medical domain, these models often fall short \cite{lehman2023we,thirunavukarasu2023large,perez2025exploring}. This performance gap requires either the development of dedicated domain-specific models, or the application of knowledge transfer techniques to adapt general-purpose models to the desired domain \cite{lehman2023we}.

In recent years, significant efforts have been made to develop medical domain-specific \gls{llm} \cite{nazi2024large} that achieve increasingly good performance on various medical tasks. In contrast, the development of medical vision models was more focused on multimodal training and the use of text supervision leveraging \gls{llm} \cite{huang2021gloria,wang2022medclip}. Nevertheless, image–text alignment may come at the cost of representation collapse, where intraclass variations may not be preserved \cite{liang2022mind}. Interestingly, recent findings suggest that pre-trained general vision domain models actually work better for certain medical tasks, such as anomaly detection \cite{lagogiannis2023unsupervised}. This unexpected performance underscores the need for specialised unimodal large medical vision models to be capable of learning transferable medical features, as recent work indicates that training unimodal medical visual representations can outperform language supervision \cite{perez2025exploring}. However, unimodal training removes the capability of multimodal models to do zero-shot classification, which is crucial for the prediction of rare and unknown diseases.

Recently, Maiorca et al. \cite{maiorca2024} introduced the concept of semantic alignment, enabling model stitching by directly translating between different latent spaces when exposed to semantically similar information. Model stitching refers to the process of combining models that have not been explicitly trained together, without requiring retraining. For example, an encoder can be paired with a decoder trained on the representation from a different encoder architecture representation, or a text encoder can be combined with a decoder trained on a visual representation.
They demonstrate the versatility of the methods in stitching models trained on diverse datasets, modalities (text and vision) and architectures, evaluating their effectiveness for classification and reconstruction tasks. Nonetheless, their experimentations are based on general-purpose models, which may not account for the specialised domain knowledge required in domain-specific tasks. This raises the question of whether it is possible to stitch medical-specific models with general models while keeping the domain-specific knowledge to handle the nuances of medical tasks.

The contributions of this work are twofold. First, we investigate whether semantic alignment between general and medical-specific models is feasible. We conduct a thorough study between a large set of general and medical domain-specific models on several public Chest X-ray datasets, including multimodal datasets with associated radiology reports. We show that aligning general models to medical-specific ones across model architectures improves standard classification task performance, illustrating the transfer of specialised medical knowledge into general-purpose AI models. Secondly, we propose a new zero-shot classification method for unimodal vision encoders that leverages semantic alignment across modalities (between image and text). Unlike the standard zero-shot classification approach, our method does not require multimodal training. We demonstrate that our approach surpasses general multimodal models' performance levels and achieves performance levels comparable to the medical multimodal model.



\section{Related Work}

Model stitching aims to combine different neural network components into a unified model. Early approaches trained a stitching layer to connect different parts of neural networks \cite{lenc2015understanding,csiszarik2021similarity}, while later work designed reusable network components that avoid a dedicated stitching layer \cite{gygli2021towards,yaman2023learning}. More recently, relative representation methods enable zero-shot model stitching by projecting model embeddings into a shared space \cite{moschellarelative}. Similar to semantic alignment, direct alignment of representational spaces via linear mapping has been explored \cite{lahner2024direct}, eliminating the need for a shared space, trainable stitching layers or predefined reusable components and allowing a more flexible and generalisable model stitching approach.

\begin{figure}[t]
    \centering
    \includegraphics[width=0.9\linewidth]{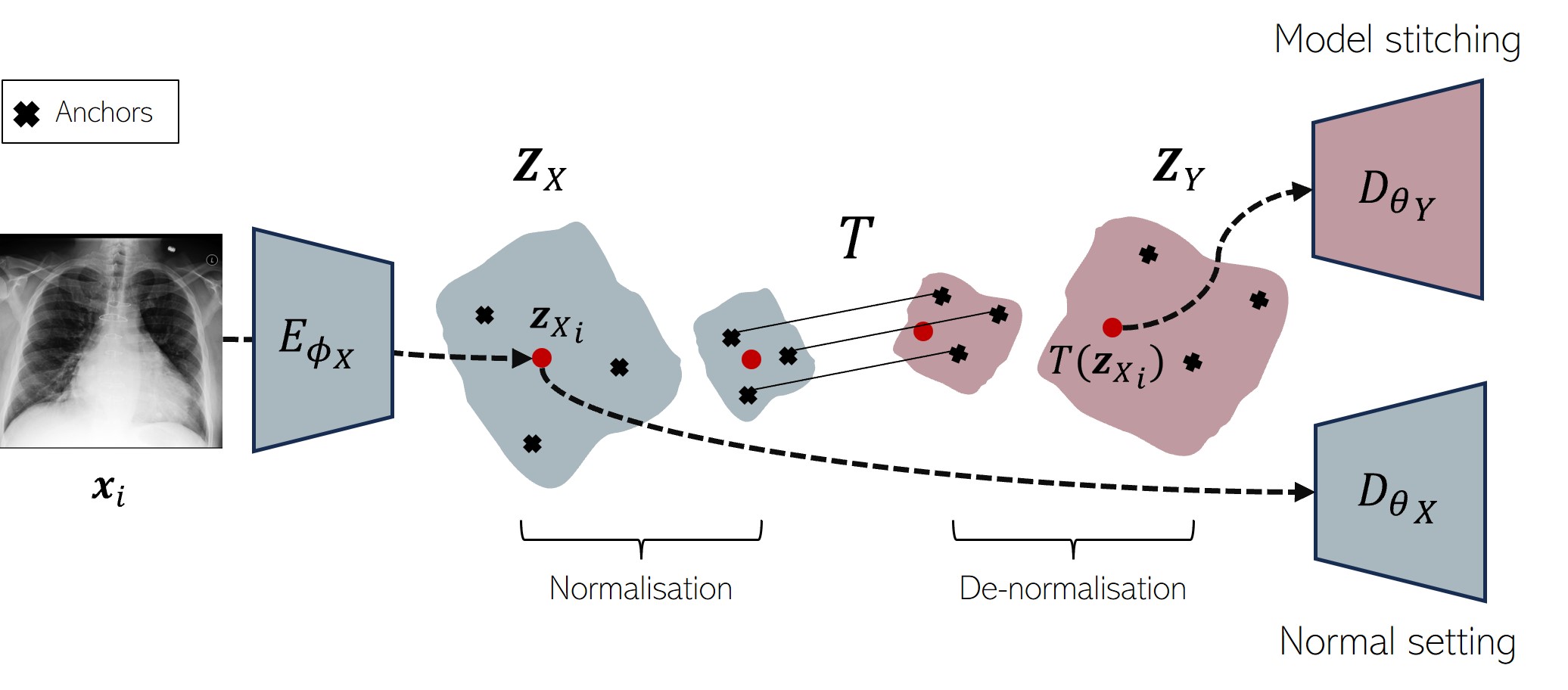}
    \caption{Illustration of the semantic alignment process, which enables model stitching between encoders from different architecture or modalities via the estimation on the anchors of a transformation $\mathcal{T}$, that is at most affine.}
    \label{fig:overview}
\end{figure}

\section{Methods}

\subsection{Semantic alignment}
\label{sec:stitching}
Considering two latent spaces $\mathbf{X} \in \mathbb{R}^{N \times d_1}$ and $\mathbf{Y} \in \mathbb{R}^{N \times d_2}$ of N samples and of dimensions $d_1$ and $d_2$ respectively, we assume access to two subsets of parallel anchors $\mathcal{A}_{\mathcal{X}} \subset \mathcal{X}$ and $\mathcal{A}_{\mathcal{Y}} \subset \mathcal{Y}$, with $\mathcal{X}$ and $\mathcal{Y}$ being data distributions, which have a semantic correspondence $\Gamma :\mathcal{A}_{\mathcal{X}} \xrightarrow{} \mathcal{A}_{\mathcal{Y}}$. Here, semantic correspondence implies that each anchor in the first set corresponds to the same high-level concept as its counterpart in the second set, e.g., a chest x-ray image and its associated radiology report. Maiorca et al. \cite{maiorca2024} demonstrated empirically that a transformation $\mathcal{T}$ can be estimated on the subset of anchors that translates between the two latent spaces $\mathbf{X}$ and $\mathbf{Y}$: $\mathbf{Y} = \mathcal{T}(\mathbf{X})$ and enables zero-shot model stitching, i.e., combining parts of different models across different trainings settings, architectures and modalities without retraining. This process, entitled semantic alignment and illustrated Fig. \ref{fig:overview}, leverages the semantic correspondence between the anchors and consists of two steps: pre-processing and estimating the transformation $\mathcal{T}$. 


\textbf{Pre-processing}: We adopt the pre-processing procedure outlined in \cite{maiorca2024}. In the case of two spaces having different dimensionalities ($d_1 \neq d_2$), we zero-pad the smaller space to align its dimensionality with the larger one while preserving its underlying structure \cite{williams2021}. Furthermore, we standardise each feature in the encoding to have zero mean and unit variance (standard scaling). The statistics for this scaling are calculated solely from the anchor sets in both the source and target spaces, enabling the required de-normalization.

\textbf{Estimating} $\mathcal{T}$: The core assumption of semantic alignment, as defined in \cite{maiorca2024}, is that $\mathcal{T}$ is at most an affine transformation $\mathcal{T} = \mathbf{R}\mathbf{X} + \mathbf{b}$. This transformation, referred to as \texttt{affine} in the experiments, is optimised using gradient descent and compared to other types of transformations, which are derived by progressively imposing additional constraints on this general form: \texttt{linear} is obtained by removing the bias term $\mathbf{b}$ and optimised using the Least Squares method; \texttt{l-ortho} is constrained to be orthogonal, ensuring that it encodes an isometry. This solution is derived by applying Singular Value Decomposition (SVD) to the $\mathbf{R}$ matrix obtained from the linear transformation \cite{xing2015}; and \texttt{ortho} corresponds to the optimal orthogonal $\mathbf{R}$, computed by Procrustes analysis \cite{gower1975}. 

\textbf{Model stitching}: Given two encoders $E_{\phi_X}$ and $E_{\phi_Y}$, model stitching consists of encoding a sample $\mathbf{x}_i$ using $E_{\phi_X}$, transforming its representation via the estimated translation $\mathcal{T}$ to align semantically with the embedding space of $E_{\phi_Y}$, and subsequently decoding it using a decoder trained on the representation from $E_{\phi_Y}$ (illustrated in Fig. \ref{fig:overview}). The semantic alignment transformation $\mathcal{T}$  is calculated a priori using an anchor set derived from the training data. We evaluated the model stitching in a cross-architecture setting, where models operating within the same modality, such as two vision encoders being stitched together. In this case, each anchor pair consists of the same sample $\mathbf{x}_i$ encoded separately encoded by two different models. 

\subsection{Unimodal Zero-shot classification}

\begin{figure}[h]
    \centering
    \includegraphics[width=0.7\linewidth]{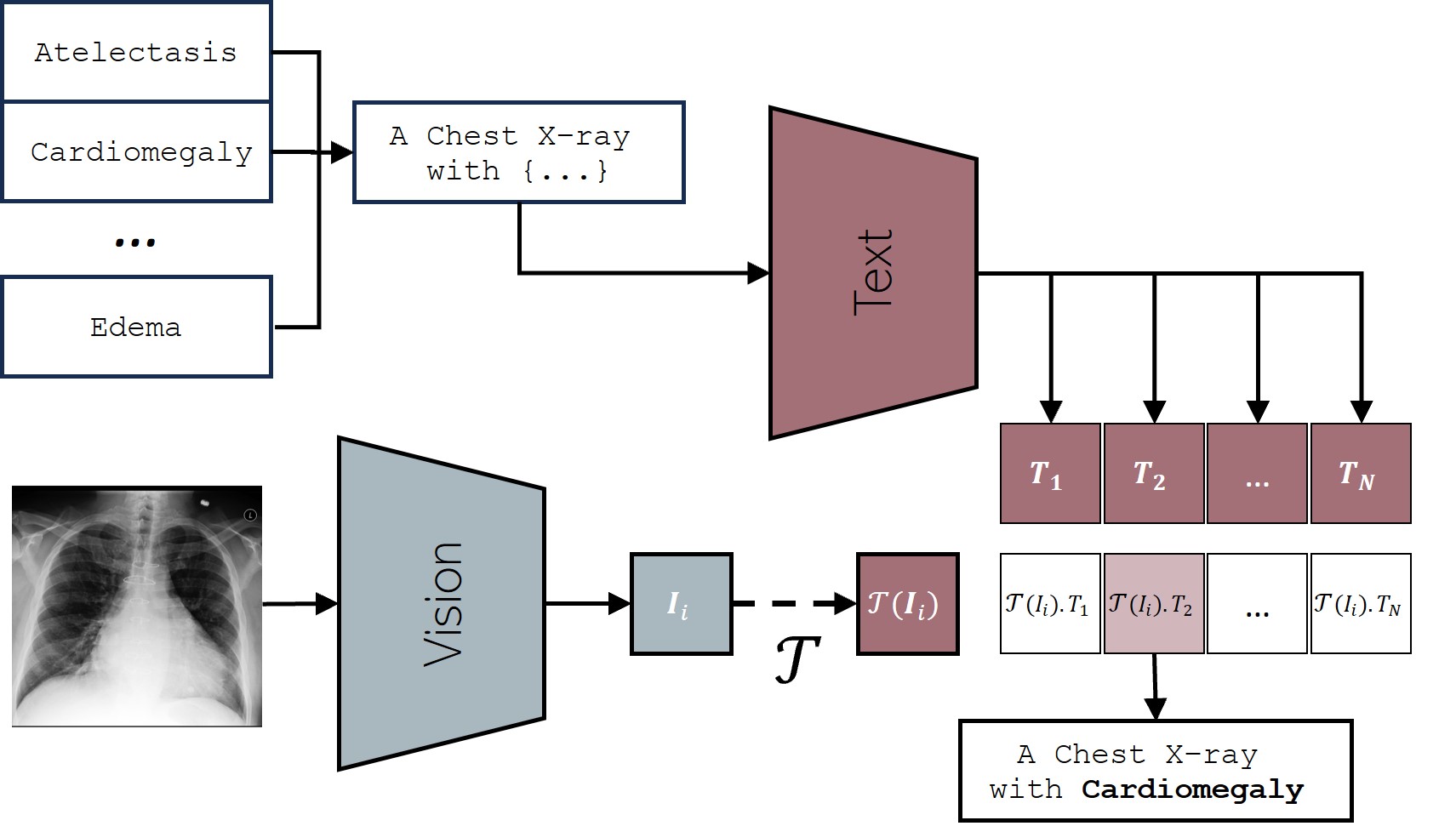} 
    
    \caption{Illustration of the proposed unimodal zero-shot classification method, which transforms image embeddings into the text space and computes similarity scores entirely between text embeddings.}
    \label{fig:zero}
\end{figure}

We propose a novel zero-shot classification paradigm that leverages semantic alignment to enable zero classification for unimodal vision encoders. In this context, semantic alignment is applied cross-modality by pairing a vision encoder with a text decoder, where each anchor pair consists of an image with its corresponding text label or description. In the standard approach, the feature embedding of an image, denoted $\mathbf{I}_i$, along with the text embeddings $\mathbf{T} = \{\mathbf{T}_i$, $i \in [1,N] \}$ for the possible $N$ classes are computed. The class with the highest cosine similarity score between the image embedding the text embeddings determines the predicted class.


In our proposed method, illustrated in Fig.~\ref{fig:zero}, after embedding the image, we transfer its embeddings to the text space using semantic alignment $\mathcal{T}(\mathbf{I}_i)$ in the cross-modality setting and compute the similarity score between the transferred embedding and the class embeddings $\mathcal{T}(I_i)$. As for the standard setting, a normalisation of the embeddings is necessary. 

\section{Experiments and results}

\textbf{Datasets}: For evaluation, we utilised: \textbf{MIMIC-CXR} \cite{johnson2019mimic} (Custom \href{https://physionet.org/content/mimic-cxr-jpg/view-license/2.1.0/}{LICENSE}), a large dataset of chest X-ray studies with associated radiology reports, served to learn cross-modal alignment in the zero-shot experiments and evaluate text classification on the findings of the radiology report on a MIMIC-5x200 subset as in \cite{wang2022medclip}, \textbf{RSNA pneumonia} \cite{shih2019augmenting} (Custom \href{https://www.kaggle.com/competitions/rsna-pneumonia-detection-challenge/rules#7-competition-data}{LICENSE}) with binary pneumonia labels, was used for classification tasks with a balanced test split for evaluation and \textbf{VinDr-CXR} \cite{nguyen2021vindr} (Custom \href{https://physionet.org/content/vindr-cxr/view-license/1.0.0/}{LICENSE}), from which we selected six findings (Aortic enlargement, Cardiomegaly, Lung Opacity, Pleural thickening, Pulmonary fibrosis and Tuberculosis as in \cite{perez2025exploring}. A balanced subset extracted from the dataset test set is sampled for each task.

\textbf{Implementation details}: We select a diverse set of models from vision and text domains, including general-purpose and medical-specific ones. For vision, we include general models: one small vision transformer (ViT) \cite{alexey2020image}, and the base model of DINOv2 \cite{oquab2023dinov2}, as well as one medical-focused model: RAD-DINO \cite{perez2025exploring}. For text, we consider general language models: RoBERTa \cite{liu2019roberta} and Llama-2 \cite{touvron2023llama} as well as domain-specialised models (BioLORD \cite{remy2022biolord} and Meditron \cite{chen2023meditron} based on Llama2). We also used the multimodal models CLIP \cite{radford2021learning} and MedCLIP \cite{wang2022medclip} as baselines for the zero-shot experiments. For each dataset and unimodal encoder (always frozen) possibility, we trained an associated decoder using an SVM with a linear kernel. For each benchmark, we compute the average performance across all possible (encoder, decoder) combinations and runs for each test set. Each experiment is repeated 3 times with a different set of anchors, each time selected randomly from the training set. We leverage the \textit{latentis} package\footnote{\url{https://github.com/Flegyas/latentis}} for the pre-processing steps and for estimating the transformation between the latent spaces. 


\subsection{Semantic alignment cross-architecture}

In this section, we experiment with the model stitching procedure (Section \ref{sec:stitching}) in a cross-architecture setting, i.e., aligning between text models for text-related tasks and between vision models for vision classification tasks.

\begin{figure}[!t]
    \centering
    \subfloat[Text]{\includegraphics[width=0.45\linewidth]{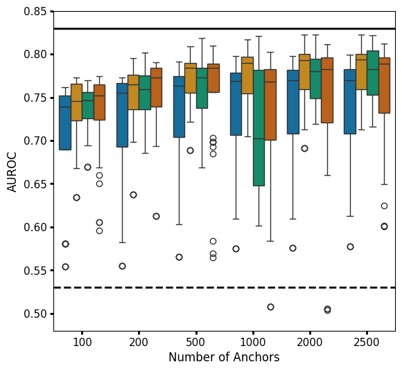}}
    \subfloat[Vision]{\includegraphics[width=0.45\linewidth]{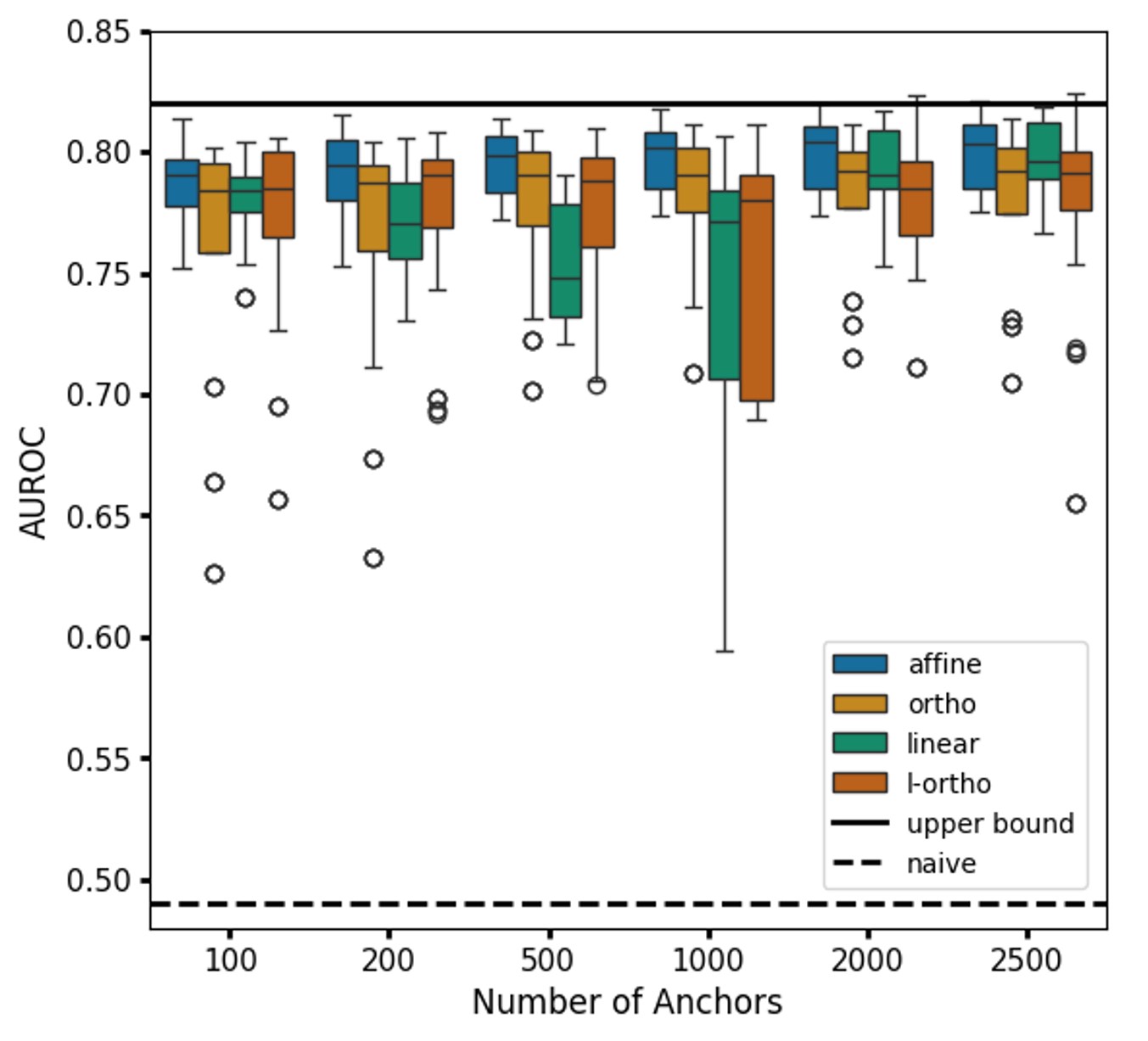}}
    \caption{Performance comparison of \texttt{affine}, \texttt{linear}, \texttt{l-ortho}, and \texttt{ortho} with varying number of anchors for (a) Text (MIMIC-CXR) and (b) Vison (RSNA) classification tasks.  \texttt{Naive} represents model stitching without semantic alignment and \texttt{upper bound} the average performance in a normal setting. Mean and standard deviation are across all possible encoders-decoders combinations and runs.}
    \label{fig:anchors}
\end{figure}

\paragraph{\textbf{Influence of $\mathcal{T}$:}} We first illustrate the impact of the number of anchors and the chosen transformation methods on MIMIC-CXR radiology reports for the text task and on RSNA pneumonia dataset for the vision one (Fig.~\ref{fig:anchors}). All alignment strategies outperform the \texttt{naive} setting where no transformation is applied, confirming that semantic alignment enables model stitching. Among the tested transformations, \texttt{ortho} performed best for text models, while \texttt{affine} yielded the highest performance for vision models. Performance improves with more anchors, particularly for text tasks, and almost reaches the \texttt{upper bound}, which corresponds to the mean performance in a normal setting (no module stitching). This shows that semantic alignment effectively enables model stitching across architectures without substantial performance degradation. For all subsequent experiments, we adopt the \texttt{ortho} transformation with 2500 anchors.

\begin{table}[h]
	\centering
    \caption{Comparison of stitching performance (AUROC) between \textit{general} (\textit{Gen}) and \textit{medical}-domains (\textit{Med}) on Text and Vision classification tasks. Alignment is applied between from \textit{Source}$\rightarrow$\textit{Target}. Source and target domain upper bound (denoted SUB and TUB respectively) are the mean performance of the models from this domain in a normal setting.  $\blacktriangle x$ \%  corresponds to the percentage of change in performance compared to the SUB coloured in green and red for increase and decrease and in grey if the difference is less than 5\%.}
	\begin{tabular}{ccllrcllr}
    \toprule
    \multicolumn{1}{c}{} & \textcolor{white}{A} & \multicolumn{3}{c}{Text} & \tinycolor{white}{A}   & \multicolumn{3}{c}{Vision} \\
    \cline{3-5} 
    \cline{7-9}
    \multicolumn{1}{c}{} & & SUB  & \multicolumn{1}{c}{Alignment} & TUB & \tinycolor{white}{a} & SUB  & \multicolumn{1}{c}{Alignment} & TUB \\
    \midrule
    \textit{Gen}$\rightarrow$\textit{Med} (Ours)
        && 0.75  & 0.80 $\pm$ 0.01 \tinycolor{teal}{$\blacktriangle$6\%} &  0.83
         &  & 0.71 & 0.69 $\pm$ 0.08 \tinycolor{gray}{$\blacktriangledown$2\%} & 0.81 \\ 

     \cline{1-5} 
     \cline{7-9}

    \textbf{Ablation study:}\\
    \textit{Gen} $\rightarrow$ \textit{Gen} 
        & & \multicolumn{1}{l}{0.75} & 0.73 $\pm$ 0.04 \tinycolor{gray}{$\blacktriangledown$3\%}  & \multicolumn{1}{r}{\multirow{3}{*}{}} 
        & & \multicolumn{1}{l}{0.71} & 0.66 $\pm$ 0.09 \tinycolor{purple}{$\blacktriangledown$7\%} & \multicolumn{1}{r}{} \\ 

    \textit{Med} $\rightarrow$ \textit{Gen} 
        & & \multicolumn{1}{l}{0.83} & 0.75 $\pm$ 0.05 \tinycolor{purple}{$\blacktriangledown$10\%} & 
         & & \multicolumn{1}{l}{0.81} & 0.70 $\pm$ 0.08 \tinycolor{purple}{$\blacktriangledown$14\%} \\ 
        
    \textit{Med} $\rightarrow$ \textit{Med} 
        & & \multicolumn{1}{l}{0.83}  & 0.82 $\pm$ 0.01 \tinycolor{gray}{$\blacktriangledown$1\%} & 
         & & \multicolumn{1}{c}{} & \multicolumn{1}{c}{-} \\ 
    \bottomrule
\end{tabular}

	\label{tab:general_specific}
\end{table}

\paragraph{\textbf{Alignment from the general to medical domain}}: We evaluated semantic alignment from general models to medical-domain models by first grouping models into their respective categories (\textit{general} or \textit{medical}) and then applying model stitching to all possible combinations between the categories. Each category has two models for each task (text or vision), with the exception of \textit{medical} vision, which only contains one (please refer to the implementation details). Table \ref{tab:general_specific} reports the classification results for text, using  MIMIC-CXR radiology reports, and vision, where the results combine those from the RSNA pneumonia task and the six VinDr-CXR tasks. Aligning \textit{general} to \textit{medical}-specific models improved performance for text classification and almost reached the target domain upper bound, indicating some medical domain-specific knowledge transfer via the alignment. In the same setting, alignment on the vision datasets decreased the performance by 2\% compared to the source domain upper bound, which represents the desired performance for model stitching. However, it outperformed the \textit{general} to \textit{general} setting (displayed in the ablation study), showing the benefit of leveraging medical-specific models, albeit with more subtle improvement. This difference between vision and text tasks in this setting can be attributed to the greater development and specialisation of text models in the medical domain compared to vision, which has only one specialised large model for this task. 

\paragraph{\textbf{Ablation study}:} We also conducted an ablation study and experimented with semantic alignment between other domain combinations (Table \ref{tab:general_specific}). We observe that alignments within the same model category, \textit{general} to \textit{general} or \textit{medical} to \textit{medical}, resulted in performance comparable to the source domain upper bound, with a maximum decrease of 7\% observed for the vision tasks in the \textit{general} to \textit{general} setting, demonstrating another empirical evidence of the capability of semantic alignment to perform model stitching. As expected, aligning \textit{medical}-specific to \textit{general} models resulted in a performance reduction of 10\% and 14\% for both text and vision, respectively.



\subsection{Unimodal zero-shot classification}

\begin{figure}[h]
    \centering
    \includegraphics[width=1\linewidth]{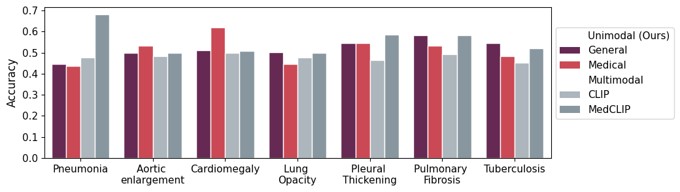}
    \caption{Comparison of the proposed unimodal zero-shot classification against multimodal methods for the RSNA pneumonia dataset and the classes of the VinDr-CXR dataset.}
	\label{fig:zero_shot}
\end{figure}

In this section, we evaluate our proposed zero-shot classification method (Fig. \ref{fig:zero}), which leverages semantic alignment cross-modality (between text and vision), on the RSNA pneumonia dataset and the six classes of the VinDr-CXR dataset. We aligned either \textit{general} or \textit{medical} vision models to the text encoder of the multimodal model CLIP \cite{radford2021learning}, which also serves as a baseline in its original multimodal form, along with MedCLIP \cite{wang2022medclip}. The transformation $\textit{T}$ is estimated a priori on the MIMIC dataset. For the positive class, prompts detailed phenotype traits (location, severity, subtype) as in \cite{wang2022medclip}. Prompts for the negatives were generated using \gls{llm} guided to generate a prompt for the opposite of the positive class, e.g., for Cardiomegaly: \texttt{"The heart appears at a normal size and shape"}. The class similarity is averaged over 10 prompts. Overall, zero-shot classification remains challenging, resulting in limited performance across all models. CLIP underperforms due to its lack of medical domain knowledge, while MedCLIP's variability is attributed to a lack of pretraining data, affecting its generalisation capabilities \cite{wang2022medclip}. Our method outperforms the general multimodal CLIP on six out of seven tasks and achieves performance comparable to the medical multimodal model MedCLIP. General unimodal vision models outperform medical-specific ones in most cases, as observed in \cite{lagogiannis2023unsupervised}, meaning that general vision encoders still learn powerful features due to the amount of pre-training data.

\section{Conclusion}

Our results demonstrate that \textit{semantic alignment can effectively distil medical knowledge without any additional training}, surpassing naive model stitching by a large margin and nearly closing the gap to domain-specific models for text-related tasks. Additionally, our zero-shot classification approach for unimodal vision encoders overcomes a major limitation of such models, offering a scalable solution to medical multimodal models. These progresses underscore the potential of lightweight adaptation strategies for medical AI, reducing reliance on costly domain-specific training while preserving flexibility and efficiency. Future work will explore optimising alignment strategies to further enhance robustness across diverse medical imaging tasks.

\bibliographystyle{splncs04}
\bibliography{ref}

\end{document}